%% file: main.tex
\documentclass[10pt,twocolumn,letterpaper]{article}

\usepackage[pagenumbers]{wacv} 

\usepackage{graphicx}
\usepackage{amsmath}
\usepackage{amssymb}
\usepackage{booktabs}
\usepackage{multirow}
\usepackage{amsmath}

\DeclareMathOperator*{\argmax}{argmax}   

\usepackage[pagebackref,breaklinks,colorlinks]{hyperref}
\usepackage[accsupp]{axessibility} 

\usepackage[capitalize]{cleveref}
\crefname{section}{Sec.}{Secs.}
\Crefname{section}{Section}{Sections}
\Crefname{table}{Table}{Tables}
\crefname{table}{Tab.}{Tabs.}

\begin{document}

\title{CorrFill: Enhancing Faithfulness in Reference-based Inpainting with Correspondence Guidance in Diffusion Models}

\author{
Kuan-Hung Liu$^1$ \quad
Cheng-Kun Yang$^2$\thanks{Now at MediaTek Inc., Taiwan.} \quad
Min-Hung Chen$^3$ \quad
Yu-Lun Liu$^1$ \quad 
Yen-Yu Lin$^1$
\\
$^1$National Yang Ming Chiao Tung University \quad
$^2$National Taiwan University \quad
$^3$NVIDIA
\\
{\url{https://corrfill.github.io/}}
}
\maketitle

\def\ourmethod{CorrFill}

\begin{abstract}
In the task of reference-based image inpainting, an additional reference image is provided to restore a damaged target image to its original state.
The advancement of diffusion models, particularly Stable Diffusion, allows for simple formulations in this task.
However, existing diffusion-based methods often lack explicit constraints on the correlation between the reference and damaged images, resulting in lower faithfulness to the reference images in the inpainting results.
In this work, we propose \ourmethod, a training-free module designed to enhance the awareness of geometric correlations between the reference and target images.
This enhancement is achieved by guiding the inpainting process with correspondence constraints estimated during inpainting, utilizing attention masking in self-attention layers and an objective function to update the input tensor according to the constraints.
Experimental results demonstrate that \ourmethod~significantly enhances the performance of multiple baseline diffusion-based methods, including state-of-the-art approaches, by emphasizing faithfulness to the reference images.
\end{abstract}

\input{1.Introduction}
\input{2.Related_Work}
\input{3.Proposed_Method}
\input{4.Experiments}
\input{5.Conclusion}

{\small
\bibliographystyle{ieee_fullname}
\bibliography{egbib}
}

\end{document}


\title{CorrFill: Enhancing Faithfulness in Reference-based Inpainting with Correspondence Guidance in Diffusion Models -- Supplementary Materials}

\author{
Kuan-Hung Liu$^1$ \quad
Cheng-Kun Yang$^2$\thanks{Now at MediaTek Inc., Taiwan.} \quad
Min-Hung Chen$^3$ \quad
Yu-Lun Liu$^1$ \quad 
Yen-Yu Lin$^1$
\\
$^1$National Yang Ming Chiao Tung University \quad
$^2$National Taiwan University \quad
$^3$NVIDIA
\\
{\url{https://corrfill.github.io/}}
}
\maketitle

\def\ourmethod{CorrFill}

In this supplementary material, we provide implementation details, showcase examples to support our proposed approaches, and present advanced analyses of the proposed method.

\section{Implementation Details}

\subsection{Implementation Details of Baselines}
For comparisons, we implement all baseline methods using the Python library Diffusers \cite{von2022diffusers}.
For Paint-by-Example\cite{yang2023PbE}, we utilize their publicly released model weights.
Side-by-side\cite{cao2024leftrefill} is essentially an inpainting base model, and we directly utilize Stable Diffusion v2 Inpainting model\footnote{https://huggingface.co/stabilityai/stable-diffusion-2-inpainting} as its implementation.
In the case of LeftRefill\cite{cao2024leftrefill}, we use the same model of Side-by-side and incorporate LeftRefill's learned prompt embedding.
We integrate IP-Adapter-Plus\cite{ye2023ipadapter} module into a Stable Diffusion Inpainting model using their released pre-trained weights.

\subsection{Implementation Details of \ourmethod}
\ourmethod~modify baseline models by substituting the attention processing function across all self-attention layers.
Correspondence estimation and attention masking are then carried out in the substituted function.
We also collect the attention maps used to optimize input latent tensor $z_t$ in the attention processing function, and the gradients are computed in the denoising main loop of the diffusion models.
Since optimizing $z_t$ requires additional memory, a gradient accumulation strategy can be employed to trade off inference time for lower memory requirements.
We conduct the experiments using an NVIDIA RTX A5000 GPU with 24GB of memory.

\subsection{Details of Dataset Sampling}
RealEstate10K is a video dataset comprising approximately 80,000 clips sourced from YouTube.
Given that the clips are recorded by cameras with stable trajectories, adjacent frames tend to exhibit high similarity.
Therefore, when selecting image pairs from RealEstate10K, we specifically consider frames that are separated by $30$ frames during the sampling process.

\subsection{Choices of Parameters}
The parameters used in the comparisons presented in the main papers are reported in Table~\ref{tab:supptable1}.
$\text{Step}_a$ and $\text{Step}_o$ represent the number of steps guided by attention masking and latent tensor optimization, respectively, out of a total of $50$ sampling steps.
$\text{Win}_a$ is the radius that determines the neighborhood of a token used in the creation of attention masks, and $\text{Win}_s$ is the radius that determines the neighborhood for the weighted average used in attention smoothing.
$\text{Str}_a$ and $\text{Str}_o$ indicate the value $v$ added to the attention mask and the weight for controlling the guidance strength of latent tensor optimization, respectively.

We selected the parameters by evaluating the subsets of our datasets.
During this evaluation, we tested various parameter settings and observed their responses in the results of different baseline methods and datasets.
The general strategy is to increase the influence of guidance for the combinations that can significantly benefit from enhanced faithfulness.

\input{Fig_Tables/Supp_table1}
\input{Fig_Tables/Supp_Figure1}
\section{Effectiveness of Proposed Components}

\subsection{Attention Smoothing}
In the quantitative ablation study presented in the main paper, the performance gains from attention smoothing are not particularly significant.
However, we provide one example demonstrating how attention smoothing serves as a crucial component in achieving accurate inpainting results in Figure~\ref{fig:suppablat}.

\subsection{Correspondence Update Policies}
In this section, we demonstrate the effectiveness of two policies including cyclic enhancement and accumulation of attention maps over timesteps.
We conduct a comparison of the correctness of the estimated correspondences against two counterparts excluding the two policies on RealEstate10K.
To estimate the correctness of the correspondences, we generate pseudo-ground truth correspondences using an image matching method \cite{shen2024gim}.
We define a correspondence with an error within the size of one token as a correct correspondence.
The counterpart that does not accumulate attention scores over time utilizes the most recently produced correspondences for guidance.
The counterpart without cyclic enhancement is guided by the correspondences computed in the first step.
The average numbers of correct correspondences during different stages of the inpainting process are illustrated in Figure~\ref{fig:numofcorr}.
The counterpart ``No acc'' fails to achieve stability, while ``No cyc.'' relies on the correspondence produced in the first step for guidance, resulting in inferior results.
The PSNR performance results for ``Ours'', ``No acc.'', and ``No cyc.'' are $27.39$dB, $27.34$dB, and $27.25$dB, respectively.
\input{Fig_Tables/Supp_Figure2}
\input{Fig_Tables/Supp_table2}

\section{Further Analysis}

\subsection{Time Efficiency}
We analyze the average execution time for the inpainting of a single input with different key components enabled, following the experimental settings described earlier. The average execution times with LeftRefill as the baseline are reported in Table~\ref{tab:supptable2}, which indicates that the latent input optimization contributes the most additional execution time within the proposed method. The increase in execution time is primarily attributed to the necessity of gradient calculation during each denoising iteration.

\subsection{Extreme Case}
Since \ourmethod~ is an improvement method designed to enhance faithfulness, it encounters certain extreme cases that challenge its performance, particularly when baseline models struggle to address them. While we previously discussed the issue of significant geometric variation in the main paper, another notable challenge for the baseline models involves large masks. The ratios of masked pixels for our generated pairs of inputs typically range from 10\% to 40\%. We find that when faced with larger masks, the inpainting results produced by LeftRefill tend to degrade to a point where \ourmethod~ is unable to enhance faithfulness effectively. Figure~\ref{fig:suppfail} illustrates this limitation of \ourmethod~ that it relies on the robustness of the baseline model. While \ourmethod~ successfully improves the results for the first row, it does not yield similar improvements for the other cases.
\input{Fig_Tables/Supp_Figure3}

\input{Fig_Tables/Supp_Figure4}
\input{Fig_Tables/Supp_Figure5}

{\small
\bibliographystyle{ieee_fullname}
\bibliography{egbib}
}

%% file: 1.Introduction.tex
\vspace{-4mm}
\section{Introduction}
\vspace{-1mm}
\input{Fig_Tables/Figure1}

Image inpainting aims to restore damaged regions of a target image.
This task is inherently ill-posed, as any plausible outcome could be considered valid.
Consequently, general image inpainting approaches are insufficient for faithfully recovering the original content of the images.
To address this issue, reference-based image inpainting introduces supplementary images, known as reference images, to guide the recovery process for damaged regions \cite{oh2019opn}.
These reference images can be photographs of the same scene with the target image, taken from different viewpoints or at different time slots.
With the guidance of reference images, it becomes more practical to restore the target image to its original state.

Denoising diffusion probabilistic models \cite{ho2020DDPM} excel as generative models, producing high-quality and diverse images \cite{Dhariwal2021beatsGAN}, and showing significant potential in reference-based inpainting\cite{yang2023PbE, ye2023ipadapter, tang2024realfill, xu2023refpaint, cao2024leftrefill}.
Existing diffusion-based methods for reference-based inpainting \cite{yang2023PbE, xu2023refpaint} focus on training or fine-tuning an image-conditioned model to fill damaged regions based on reference images.
However, they lack direct awareness of the relationships between targets and references, which is crucial for earlier approaches based on geometry matching\cite{zhou2021transfill,zhao2023geofill}.
Without this awareness, diffusion models merely conditioned on reference images fail to ensure correct reference-target geometric correlation, leading to inpainting results that do not fully adhere to the content of the references, thus losing faithfulness.
As shown in Figure~\ref{fig:teaser}, methods lacking direct reference-target awareness suffer from unwanted objects in the generated results and can lead to incorrect scene layouts, or geometry as well.

In this work, we propose \textbf{\ourmethod}, a plug-in module for reference-based inpainting diffusion models.
It estimates image correspondences between targets and references, guiding the inpainting process with these correspondences as constraints, thereby preserving reference-target geometric relationships.
To condition the inpainting process on the reference image, we stitch the reference and the target into a single image as input, allowing the self-attention layers in the diffusion models to attend across the reference and the target\cite{cao2024leftrefill}.
As demonstrated in Figure~\ref{fig:teaser}, our method collects self-attention scores at each denoising step, computes the correspondence, and guides the subsequent denoising step using this correspondence.
It is worth noting that, in the context of reference-based inpainting, the application of existing image correspondence methods is hindered by the presence of damaged regions within the target image, rendering these methods ineffective for obtaining accurate correspondences. 

Based on the observation that self-attention scores of inpainting diffusion models on the stitched image present the primitive approximations of correspondence\cite{cao2024leftrefill} even in the damaged regions, we propose utilizing the correspondences derived from attention scores as the explicit constraints for guidance.
As illustrated in Figure~\ref{fig:teaser}, the derived correspondence approximation is updated with the newly produced attention scores at each iteration, which is then used to guide the next iteration of denoising.
This cyclic interaction between correspondence approximation and inpainting enables joint improvement, progressively refining the inpainting process to achieve a more faithful result.
Furthermore, this method does not introduce additional learnable modules, making it a general strategy for reference-based inpainting diffusion models to improve the faithfulness to the reference images.

Our key contributions are summarized as follows:
First, we propose \ourmethod, a plug-in module for diffusion models, which utilizes correspondences as the explicit constraints to enhance the faithfulness to the reference images in reference-based inpainting.
Second, the cyclic enhancement strategy employed in \ourmethod~ facilitates the derivation of correspondence approximations for damaged image pairs without the need for additional training.
This is achieved through the joint refinement of the inpainting process and the correspondence approximations.
Third, our approach increases the performances of multiple diffusion model-based approaches on datasets RealEstate10K and MegaDepth. 

%% file: Fig_Tables/Figure1.tex
\begin{figure}[t!]
    \centering
    \scalebox{0.45}{\includegraphics[width=0.9\textwidth]{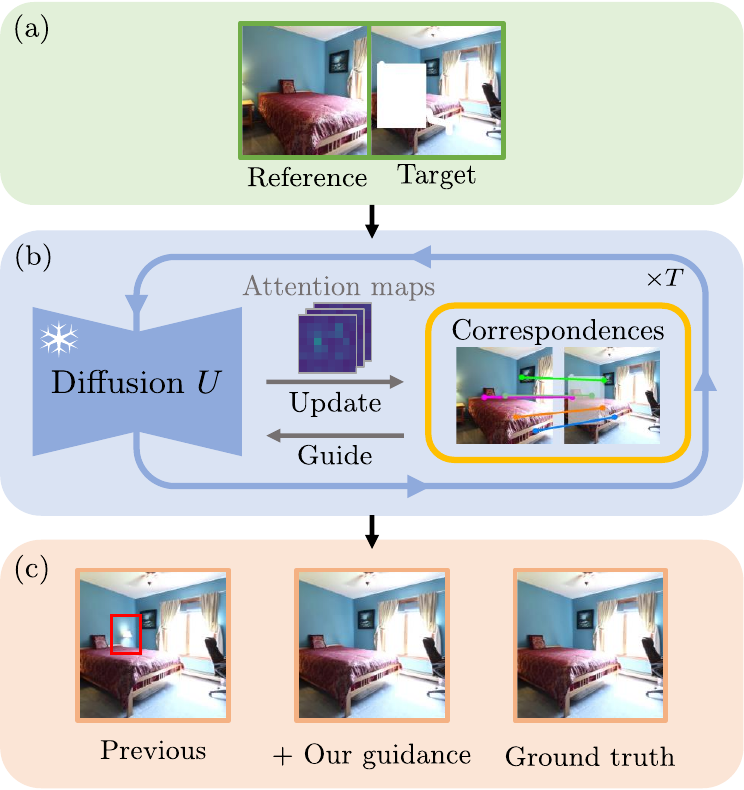}
    }
    \caption{
    \textbf{Overview.}
    (a) The reference and target images are stitched side by side, serving as inputs to the model. 
    (b) Reference-based inpainting using an inpainting fine-tuned Stable Diffusion~\cite{rombach2022LDM} that employs our training-free correspondence guidance.  
    (c) Our method captures more reliable correlations between references and targets than previous methods~\cite{cao2024leftrefill}, thereby avoiding incorrect geometry and unwanted objects. 
    }
    \label{fig:teaser}
    \vspace{-0.1in}
\end{figure}

%% file: 2.Related_Work.tex
\section{Related Work}
\input{Fig_Tables/Figure2}

\vspace{-0.05in}
\paragraph{Reference-based Image Inpainting.}
Reference-based image inpainting aims to restore images to their original states with additional knowledge provided by reference images.
Geometry-based methods\cite{zhou2021transfill, zhao2023geofill} rely on geometric estimations, resulting in complex pipelines and a tendency for error propagation, particularly when dealing with large damaged regions and insufficient overlapping areas.

The profound capability of the diffusion models\cite{ho2020DDPM} in image generation reveals their potential to perform reference-based inpainting without the need for complex pipelines.
Specifically, Stable Diffusion\cite{rombach2022LDM}, notable for its capability of high-quality text-to-image generation, spurs active development of various downstream applications, including image-conditioned generation.
Excelling in controllable generations with fine-grained guidance, image-conditioned variants of Stable Diffusion\cite{Zhang2023controlnet,Mou2024T2I,ye2023ipadapter} demonstrate significant potential for the reference-based image inpainting task.
For instance, \cite{yang2023PbE, ye2023ipadapter, tang2024realfill, cao2024leftrefill} address reference-based inpainting in an end-to-end manner based on Stable Diffusion. 

Yang~\etal~\cite{yang2023PbE} retrain a Stable Diffusion model to condition the generation process on the CLIP embedding\cite{radford2021clip} of the reference image.
LeftRefill\cite{cao2024leftrefill} employs prompt tuning techniques on a pre-trained text-to-image model, Stable Diffusion Inpainting, to avoid exhaustive training of the entire diffusion model.
This method enables a text-to-image model to work with the image condition by concatenating the reference image and the target image side by side.
Consequently, it allows mutual attention across the reference and target images via self-attention inside the diffusion U-Net.

By adopting diffusion models conditioned on reference images, these approaches circumvent the need for complex pipelines.
However, merely training a diffusion model to generate with the reference condition is insufficient to capture the correct reference-target correlation, potentially leading to inconsistent results with the reference image.
On the contrary, our \ourmethod~achieves faithful inpainting by guiding the generation process through the correspondences between reference and target images.

\vspace{-0.15in}
\paragraph{Diffusion Models Reliable Generation.}
A series of work\cite{manukyan2023hdpainter, epstein2024selfguidcontrol, chen2024layoutcontrol, chefer2023attendnexcite, feng2023structureddiffusionguidance, balaji2023ediffi, rassin2024linguisticbinding, singh2023highfidality} enhances the controllability of pre-trained text-guided diffusion models by text prompts, semantic maps, layouts, or other factors, without model re-training or fine-tuning.
For instance, Balaji~\etal~\cite{balaji2023ediffi} control the locations of generated objects by introducing cross-attention masks based on the semantic masks and corresponding text tokens, thereby encouraging semantic attributes appearing at specified image patches.
Manukyan~\etal~\cite{manukyan2023hdpainter} propose an inpainting approach that ensures faithfulness to the text prompt by reducing the self-attention scores of image tokens unrelated to the text prompt.
Furthermore, they design an objective function to enhance the effect of the text prompt on the damaged regions by optimizing the latent input of the model using the gradients of the objective function.
Without additional learnable tokens or modules, our proposed \ourmethod~incorporates correspondence into the inpainting process using attention masks and latent tensor optimization, enabling training-free fine-grained control of diffusion models.

\vspace{-0.15in}
\paragraph{Diffusion Models and Correspondence.}
The powerful image priors of pre-trained diffusion models make them foundational models for numerous applications beyond image generation, such as image correspondence.
Luo~\etal~\cite{luo2023hyperfeatures} calculate semantic correspondence by aggregating Stable Diffusion features across different network layers and diffusion timesteps using a lightweight network.
Zhang~\etal~\cite{Zhang2023zeroshot} demonstrates the capability of zero-shot semantic correspondence inherent in Stable Diffusion features.
They combine features from two vision foundation models, DINOv2 \cite{Oquab2024DINOV2} and Stable Diffusion, and perform nearest neighbor searches based on the features to establish semantic correspondence.
While these approaches estimate correspondence using features of intact images, we fully explore the potential of correspondence estimation with damaged inputs using pre-trained inpainting diffusion models. 

%% file: Fig_Tables/Figure2.tex
\begin{figure*}[t!]
    \centering
    \includegraphics[width=0.95\textwidth]{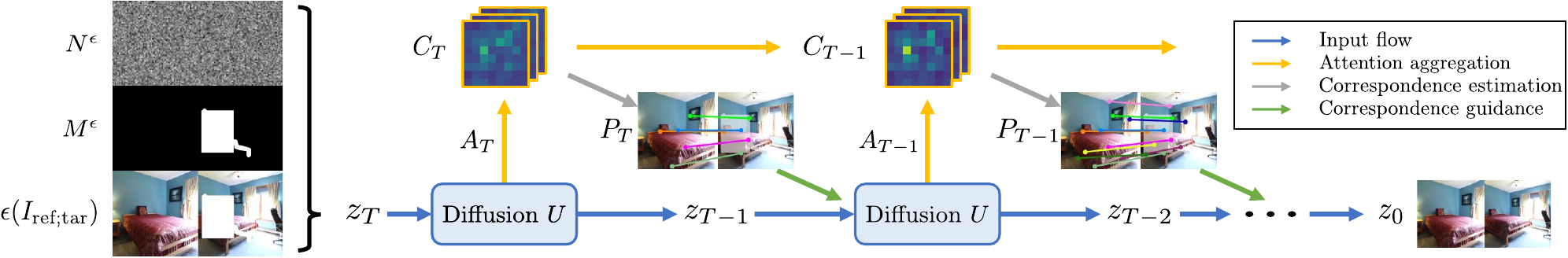}
    \caption{
    \textbf{Approach overview. }
    \ourmethod~jointly guides the inpainting and refines the estimated correspondences at each denoising step. 
    The noise tensor $N^\epsilon$, the downscaled mask $M^\epsilon$ and the encoded stitched image $\epsilon(I_{ref;tar})$ are concatenated into input latent tensor $z_T$. 
    For each denoising step, the self-attention scores from the diffusion model are aggregated into a matching map $C_t$, and the correspondence $P_t$ are computed from $C_t$, where $P_t$ are used to guide the subsequent denoising step. 
    For visual clarity, we use the real images to picture $\epsilon (I_{\text{ref;tar}})$ and $z_0$. 
    }
    \label{fig:method}
    \vspace{-0.1in}
\end{figure*}

%% file: 3.Proposed_Method.tex
\section{Proposed Method}\label{section:method}

This section presents the proposed method, \ourmethod, a correspondence-guided module for reference-based inpainting.
Firstly, a method overview is provided.
Then, we elaborate correspondence construction and refinement based on the reference-target attention scores.
Finally, we present a joint process where correspondence estimation and guided inpainting are alternately performed to facilitate each other. 

\subsection{Overview}\label{section:Overview}

Reference-based image inpainting involves a reference image $I_{\text{ref}} \in \mathbb{R}^{h \times w \times 3}$ and a target image $I_{\text{tar}} \in \mathbb{R}^{h \times w \times 3}$ with damaged regions indicated by a binary mask $M \in \{0,1\}^{h \times w}$.
As depicted in Figure~\ref{fig:method}, our proposed method aims to restore the damaged regions of $I_{\text{tar}}$ by referring to $I_{\text{ref}}$.

For ease of cross-image attention, we follow the practice used in~\cite{cao2024leftrefill} and horizontally stitch the reference and target images to yield $I_{\text{ref;tar}} \in \mathbb{R}^{h \times 2w \times 3}$.
\ourmethod~is developed based on pre-trained latent diffusion models~\cite{rombach2022LDM}.
To work in the latent space, the stitched image is encoded into $\epsilon(I_{\text{ref;tar}}) \in \mathbb{R}^{h' \times 2w' \times d}$, where $\epsilon(\cdot)$ is a variational autoencoder~\cite{kingma2022vae} and $d$ is the dimension of the latent space.
The image latent $\epsilon(I_{\text{ref;tar}})$ is then concatenated with the noise latent $N^\epsilon \in \mathbb{R}^{h' \times 2w' \times d}$ and the resized input mask $M^\epsilon \in \{0,1\}^{h' \times 2w'}$, forming the input latent tensor $z_T \in \mathbb{R}^{h' \times 2w' \times (2d + 1)}$ to the diffusion module. 

For each denoising step $t$, it is carried out by a diffusion U-Net $U$, which takes the latent tensor $z_t$ and correspondence $P_{t+1} \in [0,1]^{h' \times w' \times 2}$ computed in the previous step as input and produces $z_{t-1}$ via noise estimation.
To compute correspondence, we utilize the self-attention maps produced in the denoising process.
During denoising, the self-attention map $A_t \in \mathbb{R}^{(h' \times 2w') \times (h' \times 2w')}$ is computed and represents the patch-wise similarity in the stitched image $I_{\text{ref;tar}}$ at step $t$.
We compile a matching map $C_t \in \mathbb{R}^{h' \times w' \times h' \times w'}$ to record the consensus on patch-wise similarities across the reference and target images of all attention maps.
Namely, $C_t(i,j,\hat{i},\hat{j})$ denotes the matching degree between patch $(i,j)$ in the target and patch $(\hat{i},\hat{j})$ in the reference.
To aggregate information through the denoising process and stabilize the matching maps, $C_t$ is estimated by considering both $C_{t+1}$ and $A_t$.
We further apply the geometric constraints to $C_t$ to construct correspondence $P_t \in [0,1]^{h' \times w' \times 2}$, where $P_t(i,j)$ is the corresponding normalized coordinate in the reference of patch $(i,j)$ in the target.
The correspondence $P_t$ serves as the input and can facilitate denoising and inpainting in the next step $t-1$.

\subsection{Attention-Consensus Correspondence}\label{section:Attention}

With correspondence guidance, inpainting models can identify the most relevant parts to fill damaged regions, while avoiding interference from irrelevant parts.
However, existing correspondence estimation approaches cannot find correspondences inside the damaged region.
Inspired by semantic correspondence estimation using pre-trained diffusion models~\cite{luo2023hyperfeatures,Zhang2023zeroshot,Zhang2024Telling}, we explore the capability of generalizing an inpainting diffusion model to joint correspondence estimation and image inpainting.
Unlike methods~\cite{Zhang2023zeroshot,Zhang2024Telling} using nearest neighbor search with diffusion features, we take self-attention scores as similarity matrices so that these scores can serve as the common domain for both correspondence estimation and image inpainting.

\input{Fig_Tables/Figure3}

As shown in Figure~\ref{fig:firststep}, the self-attention scores present the correlation between references and targets even in the early generation stages.
However, the attention map from a single attention layer is often less informative. To address this, we aggregate attention maps through accumulation across different layers.
Specifically, we rescale averaged attention maps at different layers to a common size of $(h' \times 2w' \times h' \times 2w')$ and sum them up, resulting in aggregated attention map $A_t$.
Since correspondence is established across the reference and target images, we consider only the parts of self-attention scores where queries are from the target and key-value pairs are from the reference.
Therefore, the target-to-reference attention map $A_t^{\text{tar2ref}}\in \mathbb{R}^{h' \times w' \times h' \times w'}$, a submatrix of $A_t$, is extracted accordingly.

To calculate correspondence, we compute the matching map $C_t$ by merging all aggregated attention maps until the current timestep, \ie, $C_t = \sum_{\tau=t}^{T} A_{\tau}^{\text{tar2ref}}$.
The reason we choose to calculate correspondences using consensus of the aggregated attention scores from multiple layers and timesteps is to eliminate the individual biases in certain layers and timesteps.

With the matching map $C_t$, the correspondence $P_t(i,j)$ for target token $(i, j)$ is presented as the corresponding reference token and is determined via
\begin{equation}\label{equation:P}
    P_t(i,j) = \argmax_{(\hat{i},\hat{j})} C_t(i,j, \hat{i},\hat{j}),
\end{equation}
where $(i,j)$ and $(\hat{i},\hat{j})$ are the coordinates of the target and reference tokens, respectively. 

\subsection{Correspondence Refinement}\label{section:Enhancement}

As the self-attention mechanism is essential to propagating reference content to the damaged regions in the target, target query tokens attending to irrelevant reference tokens typically lead to incorrect inpainting results.
Since the preliminary correspondences $P_t$ are established by referring to merely individual reference-target token pairs, they are not stable.
Guiding the inpainting process solely on these correspondences fails to prevent the target tokens from attending to irrelevant tokens.
To this end, we propose a correspondence refining strategy, including \textit{filtering} and \textit{smoothing}, to eliminate the inaccurate correspondence in $P_t$.

\vspace{-0.15in}
\paragraph{Correspondence Filtering.}
Given that the effective correspondences only reside in the overlapping areas of the reference and target images, it is clear that not every target token has a corresponding reference token.
We observe that the target tokens not located in the overlapping regions tend to exhibit strong attention to certain reference tokens.
We define these strongly attended but irrelevant reference tokens as \textit{dominant tokens}.
They need to be removed from correspondence constraints to avoid wrong feature propagation.
Dominant tokens are identified by the presence of strong attention from diverse target tokens in $P_t$.
In practice, we consider reference tokens with more than a certain number of corresponding target tokens as dominant, and their associated correspondences are probably outliers and, therefore, are excluded from $P_t$.
We empirically set the threshold to four tokens and observe that this parameter is insensitive to the experimental results.
Additionally, we notice that some target tokens within the overlapping regions are also affected by the dominant tokens, resulting in incorrect inpainting results.
Hence, we save these excluded outlier correspondences as $P_t^o$, which are used to mitigate the adverse effects they caused through guidance.

\vspace{-0.15in}
\paragraph{Correspondence Smoothing.}
We introduce a smoothing mechanism based on the observation that when an incorrect inpainting result is present, a portion of target tokens at the center of the masked area (\ie, the damaged region) exhibit incorrect correspondences.
Conversely, their surrounding tokens, located around the edges of the mask, give more accurate correspondences and demonstrate attention scores consistent across different attention layers and timesteps.
Therefore, we employ neighborhood weighted averages for smoothing correspondence, which corrects misleading correspondence, aiming to alleviate the presence of unwanted objects and incorrect geometry.

To calculate neighborhood weighted averages on the correspondence, we create a displacement matrix $D_t \in \mathbb{R}^{h' \times w'}$ indicating the differences between each target token and its corresponding reference tokens in coordinate, \ie, $D_t(i,j) = P_t(i,j) - (i,j)$.
Next, we construct the consensus matrix $W_t \in \mathbb{R}^{h' \times w'}$ by assigning the matching score $C_t(i,j,P_t(i,j))$ to $W_t(i,j)$ for target token $(i,j)$, whose corresponding reference token is $P_t(i,j)$.
For outlier correspondences $P_t^o$, we set their consensus value to zero, and therefore they are ignored during the smoothing process.
The neighborhood weighted average of $D_t$ is then calculated using $W_t$ as weights as follows: 
\begin{equation}\label{equation:P'}
    D_t^\ast(i,j) = \frac{1}{|W_t(i,j)|} \sum_{(\hat{i},\hat{j}) \in \mathcal{N}(i,j)} D_t(\hat{i},\hat{j}) \cdot W_t(\hat{i},\hat{j}),
\end{equation}
where $\mathcal{N}(i,j)$ is the set of neighborhood tokens of token $(i,j)$, and $|W_t(i,j)| = \sum_{(\hat{i},\hat{j}) \in \mathcal{N}(i,j)}W_t(\hat{i},\hat{j})$.
In this formulation, we can propagate more accurate correspondences with higher degrees of consensus to those tokens of incorrect correspondences in the form of displacements, and the smoothed displacements are converted back to correspondences through $P_t^\ast(i,j) = D_t^\ast(i,j) + (i,j)$.
The value of the smoothed correspondence $P_t^\ast$ is then assigned back to the original correspondence: $P_t^\ast \rightarrow P_t$.

\input{Fig_Tables/Figure4}

\subsection{Cyclic Enhancement}\label{section:Cyclic}

By applying correspondence constraints to the denoising process, \ourmethod~establishes a cyclic enhancement that jointly improves the correspondence and inpainting processes at each iteration, progressively guiding the generation toward a faithful result.
Figure~\ref{fig:step} illustrates one cycle of the cyclic enhancement during a denoising step.
Given the estimated correspondence $P_{t+1}$ from the previous step, we guide the denoising process of the diffusion model by employing attention masks $m_t$ across all self-attention layers and further enhancing the input latent $z_t$ with an objective function $S$.
The produced attention map $A_t^{\text{tar2ref}}$ is then used to enhance the estimated correspondence $P_{t+1}$ to $P_t$ for the next step through updating the matching map $C_t$.

\vspace{-0.15in}
\paragraph{Attention Masking.}
To integrate correspondence constraints into the diffusion model, we employ attention masks within each self-attention layer.
These attention masks are incorporated into the affinity matrix to modulate the influence of different value tokens.

The attention mechanism evaluates the contribution of value tokens through the affinity matrix, expressed as $QK^{\top}/\sqrt{d_a} \in \mathbb{R}^{(h' \times 2w') \times (h' \times 2w')}$, where $Q$ and $K$ are query and key token vectors, respectively, and $d_a$ is the embedding dimension.
For ease of discussion, we focus on operations conducted at a scale of $1/8$, while these operations are consistent across all attention layers, regardless of scale.
The attention mask $m_t \in \mathbb{R}^{(h' \times 2w') \times (h' \times 2w')}$ adjusts the contribution of value tokens by adding either negative or positive values to the affinity matrix, resulting in the modified attentions: $(QK^T + m_t)/\sqrt{d_a} \in \mathbb{R}^{(h' \times 2w') \times (h' \times 2w')}$.

We represent the attention mask in the shape of ${h' \times 2w' \times h' \times 2w'}$, which preserves the spatial context for both the queries and keys.
We define a slice of the attention mask for a token $(i,j)$ as $m_t^{ij} \in \mathbb{R}^{h' \times 2w'}$, denoting the part where the dot product between the query $(i,j)$ and all keys occurs.
The attention masks are composed according to the estimated correspondence $P_{t+1}$ from the previous denoising step.
For a target token $(i,j)$ whose correspondence is not an outlier, the element in the slice $m_t^{ij}$ is defined by
\begin{equation}\label{equation:inlier}
{
    m_t^{ij}(\hat{i},\hat{j})=
        \begin{cases}
            v,& \text{if } (\hat{i},\hat{j}) \in \mathcal{N}(P_{t+1}(i,j)),\\
            -\infty,& \text{if } (\hat{i},\hat{j}) \in \mathcal{R} - \mathcal{N}(P_{t+1}(i,j)),\\
            0,& \text{otherwise},\\
        \end{cases}
    }
\end{equation}
where $v$ represents a small positive number, $\mathcal{N}(P_{t+1}(i,j))$ denotes the neighboring tokens of $P_{t+1}(i,j)$, and $\mathcal{R}$ refers to the set of all reference tokens.
When the attention mask is applied to a self-attention layer, this slice of the mask boosts the attention values of the corresponding areas, thereby promoting attention for the relevant tokens.
Conversely, it diminishes the attention values for other reference tokens, preventing them from being attended to. 

For outlier tokens in $P_{t+1}^o$, the values assigned to their slices are defined as follows: 
\begin{equation}\label{equation:outlier}
{
    m_t^{ij}(\hat{i},\hat{j})=
        \begin{cases}
            -\infty,& \text{if } (\hat{i},\hat{j}) \in \mathcal{R} \cap \mathcal{N}(P_{t+1}^o(i,j)),\\
            0,& \text{otherwise,}\\
        \end{cases}
    }
\end{equation}
This slice of the attention mask prevents the token $(i,j)$ from attending to the irrelevant area, which is identified by the outlier correspondences.
The remaining elements of the attention mask are assigned to $0$, thereby preserving the original attention values for those tokens. 
\input{Fig_Tables/table1}
\vspace{-0.15in}
\paragraph{Latent Tensor Optimization.}
Similar to the observations made in recent studies on reliable generation within diffusion models \cite{chen2024layoutcontrol,manukyan2023hdpainter}, we notice that solely employing attention masking is insufficient for steering inpainting towards the desired outcomes.
To address this issue, we adopt a similar strategy, utilizing the produced constraints for further guidance by optimizing the latent tensor $z_t$ with an objective function $S$.
The core concept is to optimize $z_t$ in a direction that aligns with the desired outcomes, specifically by ensuring that the attention of a token adheres to the pattern prescribed by $P_{t+1}$.

As depicted in Figure~\ref{fig:step}, we collect attention maps from all self-attention layers within $U$.
Similar to the process producing $A_t^{\text{tar2ref}}$, the attention maps are reshaped, resized, and used to extract the target-to-reference submatrix, resulting in $(A_l)_t^{\text{tar2ref}}$, where $l$ denotes the layer it is collected from.
Instead of aggregating them, we calculate their gradients of the objective function $S$ separately and update the input latent $z_t$ by gradient descent.
The objective function $S$ is defined as follows:
\begin{equation}\label{equation:objective}
\text{\footnotesize{$S((A_l)_t^{\text{tar2ref}}) = \text{BCE}(\text{Sigmoid}(\text{Norm}((A_l)_t^{\text{tar2ref}})), E(P_{t+1})),$}}
\end{equation}
where function $\text{Norm}(\cdot)$ normalize matrix $(A_l)_t^{\text{tar2ref}}$, and $\text{BCE}(\cdot)$ is the weighted binary cross-entropy to $[0,1]$
$E(\cdot)$ turns $P_{t+1}$ into a one-hot tensor of the same shape as $(A_l)_t^{\text{tar2ref}}$.
In this formulation, the input latent $z_t$ is optimized toward a direction where its attention maps are encouraged to adhere to the correspondence constraint.

\vspace{-0.15in}
\paragraph{Implementation Details.}
Our proposed \ourmethod~is developed based on Stable Diffusion v2 Inpainting\footnote{https://huggingface.co/stabilityai/stable-diffusion-2-inpainting} \cite{rombach2022LDM}, designed to work as a plug-in compatible with various diffusion models.
The image resolution for both target and reference images is set to $512 \times 512$, while the size of the latent representation for the stitched image $\epsilon(I_{\text{ref;tar}})$ is $64 \times 128$.
DDIM \cite{song2021ddim} sampling is used for efficient generation, with the number of sample steps set to $50$.
Further details, including the choices of parameters, are provided in the supplementary.

%% file: Fig_Tables/Figure3.tex
\begin{figure}[t!]
    \centering
    \includegraphics[width=0.45\textwidth]{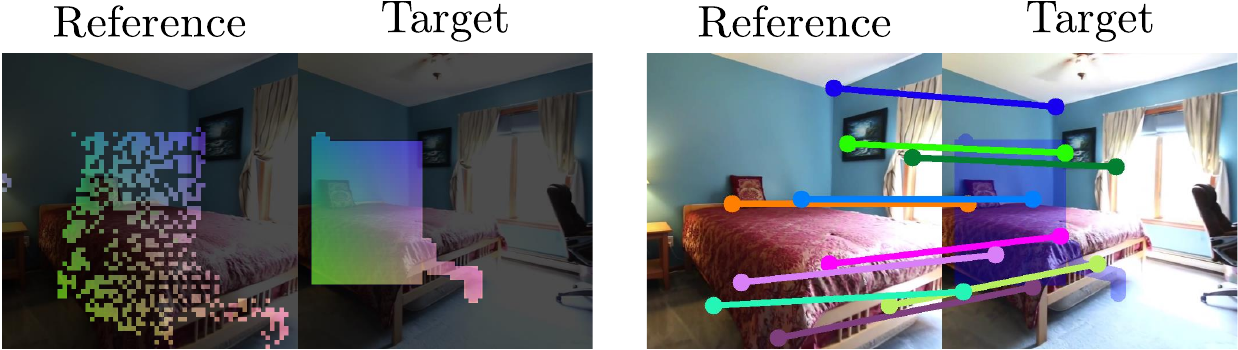}
    \vspace{-0.05in}
    \caption{
    \textbf{Correspondences in the early stage. }
    The image on the left highlights the masked regions of the target and their most attended positions in the reference, indicated by colors, at the very first denoising step. 
    The image on the right depicts a few correspondences computed at the first denoising step. 
    }
    \label{fig:firststep}
    \vspace{-0.15in}
\end{figure}

%% file: Fig_Tables/Figure4.tex
\begin{figure}[t!]
    \centering
    \includegraphics[width=0.45\textwidth]{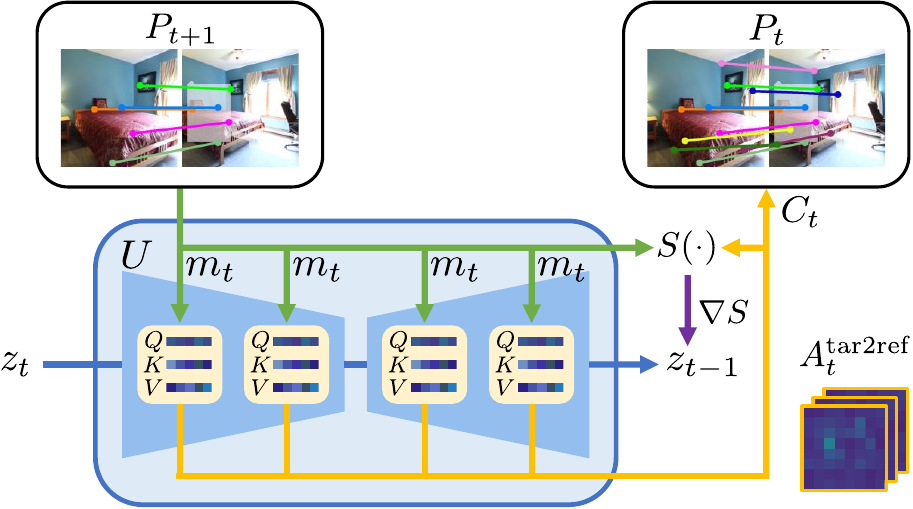}
    \vspace{-0.05in}
    \caption{
    \textbf{Correspondence guidance in the diffusion U-Net.}
    At each denoising step $t$, the denoising process is guided by the correspondences estimated in the previous step, $P_{t+1}$, through attention masking with $m_t$ and optimizing $z_t$ using the objective function $S(\cdot)$. 
    The generated attention maps $A_t^{\text{tar2ref}}$ are then employed to further refine the estimated correspondences $P_t$ by updating the matching map $C_t$. 
    }
    \label{fig:step}
    \vspace{-0.1in}
\end{figure}

%% file: Fig_Tables/table1.tex
\begin{table*}[t!]
    \centering
    \scalebox{0.8}{
        \begin{tabular}{llllllll}
            \toprule\midrule
            \multirow{2}{*}{Method}&&\multicolumn{3}{c}{RealEstate10K}&\multicolumn{3}{c}{MegaDepth}\\
            \cmidrule(r){3-5}\cmidrule(r){6-8}
            &&PSNR$\uparrow$&SSIM$\uparrow$&LPIPS$\downarrow$&PSNR$\uparrow$&SSIM$\uparrow$&LPIPS$\downarrow$\\
            \midrule
            \multirow{2}{*}{Paint-by-Example}&Baseline&20.03&0.8528&0.1379&20.48&0.8274&0.1138\\
            &+\ourmethod&21.57(+1.54)&0.8724(+0.0196)&0.1097(-0.0282)&21.02(+0.54)&0.8343(+0.0069)&0.1014(-0.0124)\\
            \midrule
            \multirow{2}{*}{IP-Adapter-Plus}&Baseline&21.26&0.8704&0.1127&21.33&0.8394&0.0989\\
            &+\ourmethod&25.10(+3.84)&0.8990(+0.0286)&0.0642(-0.0485)&22.14(+0.81)&0.8473(+0.0079&0.0838(-0.0151)\\
            \midrule
            \multirow{2}{*}{Side-by-side}&Baseline&23.32&0.8941&0.0856&22.89&0.8538&0.0850\\
            &+\ourmethod&25.81(+2.49)&0.9092(+0.0151)&0.0552(-0.0304)&23.24(+0.35)&0.8571(+0.0033)&0.0777(-0.0073)\\
            \midrule
            \multirow{2}{*}{LeftRefill}&Baseline&26.71&0.9163&0.0443&23.60&0.8649&0.0653\\
            &+\ourmethod&26.97(+0.26)&0.9175(+0.0012)&0.0427(-0.0016)&23.60&0.8649&0.0651(-0.0002)\\
            \midrule\bottomrule
        \end{tabular}
        }
    \vspace{-2mm}  
    \caption{\textbf{Quantitative results. } We demonstrate the evaluations of $4$ different baselines: Paint-by-Example\cite{yang2023PbE}, IP-Adapter-Plus\cite{ye2023ipadapter}, Side-by-side\cite{cao2024leftrefill} and LeftRefill\cite{cao2024leftrefill}, with and without the application of our \ourmethod~on the dataset RealEstate10K and MegaDepth. }
    \label{tab:table1}
    \vspace{-3mm}  
\end{table*}

%% file: 4.Experiments.tex
\section{Experiments}
\subsection{Experimental Settings}\label{section:expsettings}

\input{Fig_Tables/Figure5}
\paragraph{Datasets.}
Since there is no publicly available benchmark for this task, we follow previous works~\cite{cao2024leftrefill, zhao2023geofill} to prepare the dataset for evaluation.
We conducted experiments on subsets sampled from two datasets: RealEstate10K \cite{zhou2018realestate} and MegaDepth \cite{Li2018mega}.
We sampled $500$ pairs of references and targets from each dataset and generated the inpainting masks based on the content of image pairs. 

Before the sampling process, we cropped the images into squares and resized them to a resolution of $512 \times 512$.
To identify suitable images for experimentation with reference-based inpainting, we utilized the mid-level vision similarity metric DreamSim \cite{Fu2023dreamsim}.
Images from the same scene are considered to form an image pair if the DreamSim distance is below $0.2$.
This threshold indicates that the image pair exhibits a certain degree of similarity in both appearance and semantics, making it suitable for evaluating reference-based inpainting.
To prevent images in a pair from being overly similar, the pairs with distance below $0.1$ are discarded for RealEstate10K, for its images are likely to be overly similar compared to those in MegaDepth.

To consistently reflect the capability of reference-based inpainting, we generate inpainting masks based on content in the references and targets.
Specifically, we employ feature matching \cite{tang2022quadtree} to identify corresponding keypoints on the intact image pairs, subsequently creating masks that cover portions of these keypoints on the target images.
This approach ensures that the recovery process for damaged regions relies on the content of the reference images.
To simulate real-world applications, the masks are randomly generated, combining the shape of a rectangle with several strokes. The generated dataset will be publicly available.

\vspace{-1mm}
\paragraph{Evaluation Metrics.}
We follow the common practice \cite{zhou2021transfill,zhao2023geofill,cao2024leftrefill} and employ three evaluation metrics: Peak Signal-to-Noise Ratio (PSNR), Structural Similarity Index (SSIM) and Learned Perceptual Image Patch Similarity (LPIPS).
These metrics evaluate the similarities between restored targets and their ground truths. 

\subsection{Experimental Results}\label{section:comparison}

Our method serves as a plug-in module. We perform a thorough comparison of our method with four existing baselines in Table~\ref{tab:table1} and Figure~\ref{fig:quality}. 

\paragraph{Baselines.}
The comparison baselines achieve reference-based inpainting through various techniques.
LeftRefill \cite{cao2024leftrefill} stitches the reference and target images side by side, filling damaged regions with Stable Diffusion Inpainting \cite{rombach2022LDM}, and incorporating prompt-tuning technique to learn an optimized prompt embedding specifically for the reference-based inpainting task.
Side-by-side inpainting \cite{cao2024leftrefill} is a variant of LeftRefill, which does not employ the prompt-tuning technique.
In practice, we implement Side-by-side by providing an empty text prompt, which means that the model is not given any explicit instructions regarding our task. 

Although Paint-by-Example \cite{yang2023PbE} is designed for reference-based inpainting, its goal does not completely align with those of other baselines.
Rather than restoring damaged regions by adhering to the fine-grained details of the reference image, this method focuses on generating plausible results based solely on the semantic attributes of the reference images.
While it is conditioned on the reference images using their CLIP embeddings, it introduces an information bottleneck by considering only the global attributes, leading to a loss of fine-grained details.
IP-Adapter-Plus \cite{ye2023ipadapter} presents another approach that conditions the diffusion model on the CLIP embeddings of the reference images.
In contrast to Paint-by-Example, IP-Adapter-Plus retains fine-grained details by incorporating all spatial tokens of CLIP embeddings.

To compare against these baseline methods, we integrate \ourmethod~ into them.
For LeftRefill and Side-by-side, since they already employ the same formulation of stitched image inputs, we directly apply \ourmethod~ to their diffusion models.
For Paint-by-Example and IP-Adapter-Plus, we modify their inputs to match the stitched reference formulation and then apply our \ourmethod.

\vspace{-2mm}
\paragraph{Quantitative Results.}
We evaluate four baseline methods and their counterparts involving \ourmethod~ module.
As shown in Table~\ref{tab:table1}, our method consistently shows improvement across all baselines on RealEstate10K.

With \ourmethod, the performance of Side-by-side is elevated to $25.81$dB in PSNR, increased by $2.49$dB.
Since the model of Side-by-side is not specifically designed for reference-based inpainting, these improvements highlight the effectiveness of \ourmethod~ in enhancing the model's awareness of reference-target correlations.
While IP-Adapter-Plus preserves the appearance details in the reference images, it struggles to capture correct spatial correlations.
\ourmethod~ significantly improves its performance by $3.84$dB in PSNR through incorporating correspondence constraints.
Although the enhanced results of Paint-by-Example are inferior to those of other approaches due to the re-training of the model on a slightly different task, \ourmethod~ still enhances their performance by incorporating additional reference content.
\ourmethod~further improve the performance of LeftRefill, which is the state-of-the-art approach to the best of our knowledge.

The improvements on the challenging MegaDepth dataset are more limited due to the significant changes in viewpoints, which can lead to failures in correspondence estimation.
Additionally, the nature of the dynamic scenes may diminish the benefits gained from strictly adhering to the reference images.
Despite these challenges, \ourmethod~ still demonstrates clear improvements for Paint-by-Example, IP-Adapter-Plus, and Side-by-side across all evaluation metrics.

\paragraph{Qualitative Results.}
Figure~\ref{fig:quality} presents a quality comparison of four baseline methods and their variants integrated with \ourmethod.
In the comparison involving IP-Adapter-Plus, Side-by-side, and LeftRefill on the RealEstate10K dataset, our method effectively addresses the issues present in the baseline results, including the removal of unwanted objects in (c) and the correction of incorrect scene layouts in (a),(b), and (c).
While our method does not completely resolve all issues in Paint-by-Example, it still achieves a higher level of faithfulness in the results.
In the comparisons on MegaDepth, results exhibiting greater faithfulness can be observed when compared to most of the baselines.

\input{Fig_Tables/table2}
\subsection{Ablation Study}\label{section:ablation}

We perform an ablation study by incrementally activating different key components of \ourmethod~ to assess the impact of each component in Table~\ref{tab:table2}.
As attention masking serves as a prerequisite for outlier filtering and correspondence smoothing, our analysis reveals that the correspondence refinement strategies effectively enhance performance when applied in conjunction with attention masking.
Although the improvement brought by correspondence smoothing may seem subtle, it can be the crucial component for correcting incorrect content in certain cases, and an example is provided in the supplementary.
The optimization of the latent tensor $z_t$ further boosts performance, while also benefiting from the correspondence refinements. 

\input{Fig_Tables/Figure6}

%% file: Fig_Tables/Figure5.tex
\begin{figure*}[t!]
    \centering
    \includegraphics[width=1.0\textwidth]{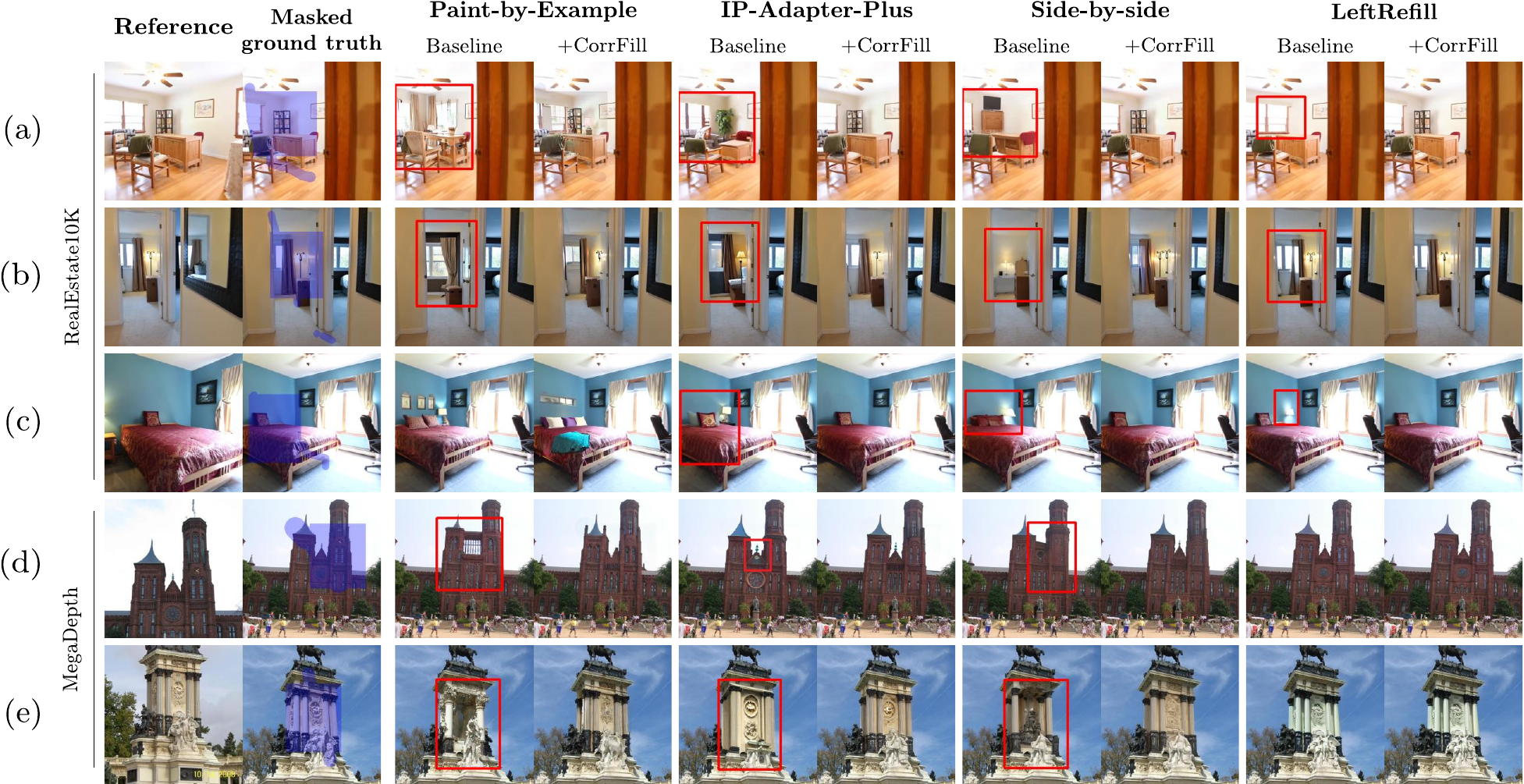}
    \caption{
    \textbf{Qualitative results. }
    We present the qualitative results with four different baselines and their counterparts integrated with our method on two datasets. 
    We highlight the problematic regions in the results of the baseline methods that our approach can effectively address by enclosing them in red boxes.
    The inpainting masks are generated based on the content in image pairs.
    }
    \label{fig:quality}
    \vspace{-0.1in}
\end{figure*}

%% file: Fig_Tables/table2.tex
\begin{table}[t]
    \centering
    \scalebox{0.7}{
        \begin{tabular}{llccc}
            \toprule\midrule
            Baselines&Module Components&PSNR$\uparrow$&SSIM$\uparrow$&LPIPS$\downarrow$\\
            \midrule
            \multirow{5}{*}{Side-by-side}&Baseline&23.32&0.8941&0.0856\\
            &+ Attention Masking&23.70&0.8995&0.0736\\
            &+ Outlier Filtering&24.36&0.9031&0.0692\\
            &+ Correspondence Smoothing&24.37&0.9030&0.0694\\
            &+ Latent $z_t$ Optimization&25.81&0.9092&0.0552\\
            \midrule\
            \multirow{5}{*}{LeftRefill}&Baseline&26.71&0.9163&0.0443\\
            &+ Attention Masking&26.45&0.9153&0.0458\\
            &+ Outlier Filtering&26.78&0.9165&0.0438\\
            &+ Correspondence Smoothing&26.79&0.9166&0.0438\\
            &+ Latent $z_t$ Optimization&26.97&0.9175&0.0427\\
            
            \midrule\bottomrule
        \end{tabular}    
        }
    \vspace{-2mm}  
    \caption{\textbf{Ablation study on key components of \ourmethod. } The ablation study on key components of \ourmethod~is conducted with Side-by-side and LeftRefill \cite{cao2024leftrefill}~ baselines, on RealEstate10K dataset. It is important to note that outlier filtering and correspondence smoothing can only be implemented when attention masking is enabled. }
    \label{tab:table2}
    \vspace{-3mm}  
\end{table}

%% file: Fig_Tables/Figure6.tex
\begin{figure}[t!]
    \centering
    \includegraphics[width=0.45\textwidth]{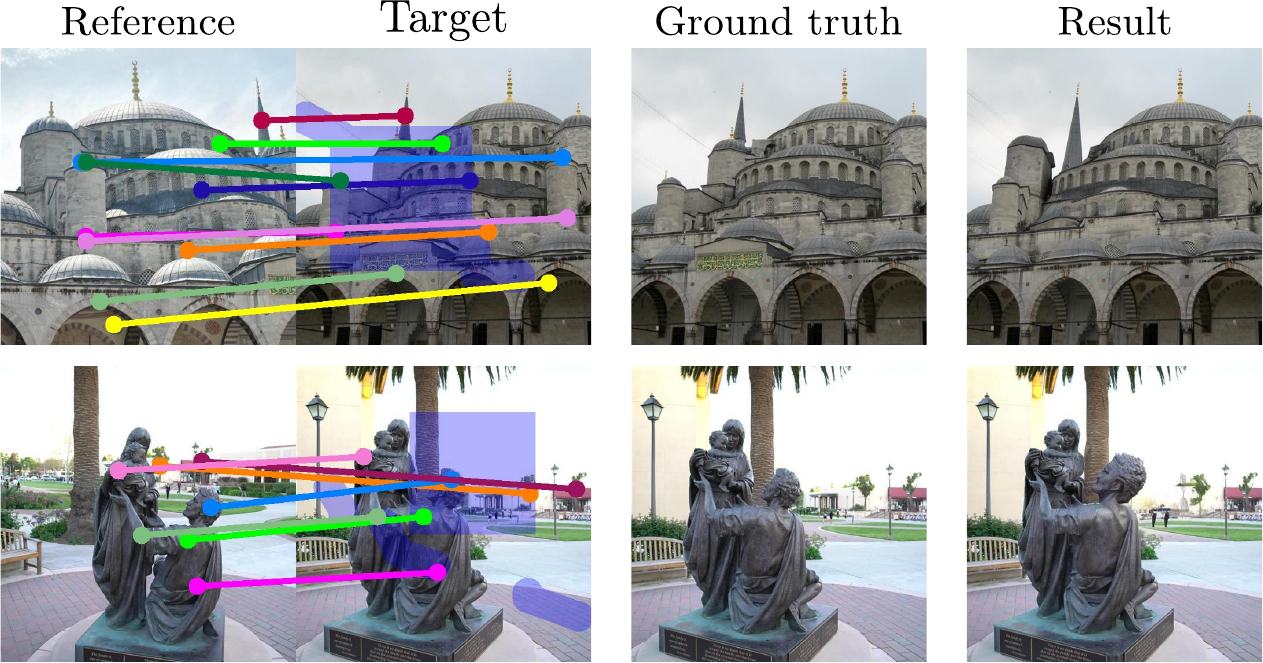}
    \caption{
    \textbf{Failure cases. }
    The image on the left depicts the estimated correspondences. 
    The image on the right shows the inpainting result of \ourmethod~integrated on LeftRefill\cite{cao2024leftrefill}.
    In the first row, the repetitive structures and complex geometry introduce ambiguity in correspondence estimation.
    In the second row, the incorrect orientation of the statue's head demonstrates that correspondence constraints in 2D space are inadequate when faced with significant changes in viewpoint due to a lack of 3D awareness.
    }
    \label{fig:failure}
    \vspace{-0.1in}
\end{figure}

%% file: 5.Conclusion.tex
\section{Conclusion}

We propose \ourmethod, a training-free module that incorporates correspondence constraints into reference-based image inpainting diffusion models.
\ourmethod~achieve higher degrees of faithfulness to the reference images in the inpainting results by guiding the inpainting process with correspondence between the reference and target images.
To perform this guidance, we exploit the capability of diffusion models to estimate correspondence during the inpainting process, and we utilize this correspondence to constrain the inpainting through self-attention masking and input latent optimization.
Experimental results demonstrate the effectiveness of \ourmethod~in enhancing reference-target correlations in the inpainting results for multiple baseline methods.
\ourmethod~improves performance across these baselines on RealEstate10K and MegaDepth datasets, pushing the limits for state-of-the-art reference-based inpainting methods.

\noindent\textbf{Limitations.}
As shown in the first row of Figure~\ref{fig:failure}, \ourmethod~fails to faithfully restore damaged regions when encountered with scenes featuring complex geometry or repetitive structures, which introduce significant ambiguity when estimating correspondences.
In the second row, although \ourmethod~ successfully captures the reference-target correlation, the incorrect orientation of the statue's head suggests that our correspondence constraints in 2D space are susceptible to the significant geometric variations of objects, which require advanced 3D prior.

\noindent\textbf{Acknowledgement.}
This work was supported in part by the National Science and Technology Council (NSTC) under grants 112-2221-E-A49-090-MY3, 111-2628-E-A49-025-MY3, and 112-2634-F-002-005. This work was funded in part by NVIDIA.

%% file: Fig_Tables/Supp_table1.tex
\begin{table*}[t!]
    \centering
    \scalebox{0.75}{
        \begin{tabular}{lcccccccc}
            \toprule\midrule
            \multirow{2}{*}{Parameter}&\multicolumn{2}{c}{Paint-by-Example}&\multicolumn{2}{c}{IP-Adapter-Plus}&\multicolumn{2}{c}{Side-by-side}&\multicolumn{2}{c}{LeftRefill}\\
            \cmidrule(r){2-3}\cmidrule(r){4-5}\cmidrule(r){6-7}\cmidrule(r){8-9}
            &RealEstate10K&MegaDepth&RealEstate10K&MegaDepth&RealEstate10K&MegaDepth&RealEstate10K&MegaDepth\\
            \midrule
            $\text{Step}_a$&50&25&25&25&50&25&5&5\\
            $\text{Step}_o$&50&50&50&50&50&50&50&5\\
            $\text{Win}_a$&4\text{(t)}&4\text{(t)}&0.3\text{(i)}&5\text{(t)}&2\text{(t)}&3\text{(t)}&0.3\text{(i)}&2\text{(t)}\\
            $\text{Win}_s$&0.4&0.4&0.2&0.2&0.2&0.2&0.05&0.2\\
            $\text{Str}_a$&1&1&1&1&1&1&1&0\\
            $\text{Str}_o$&2&0.5&2&0.5&2&0.5&0.5&0.5\\
            
            \midrule\bottomrule
        \end{tabular}
        }
    \vspace{-2mm}  
    \caption{\textbf{List of parameters. } The comparisons presented in the main papers are conducted using these parameters. 
    (t) indicates that the value refers to the number of tokens, and (i) denotes that the value is the ratio to the size of encoded images, \ie, $h'$.
    For ${\text{Win}}_s$, all the values are the ratios to the size of encoded images.}
    \label{tab:supptable1}
    \vspace{-3mm}  
\end{table*}

%% file: Fig_Tables/Supp_Figure1.tex
\begin{figure}[t!]
    \centering
    \includegraphics[width=0.45\textwidth]{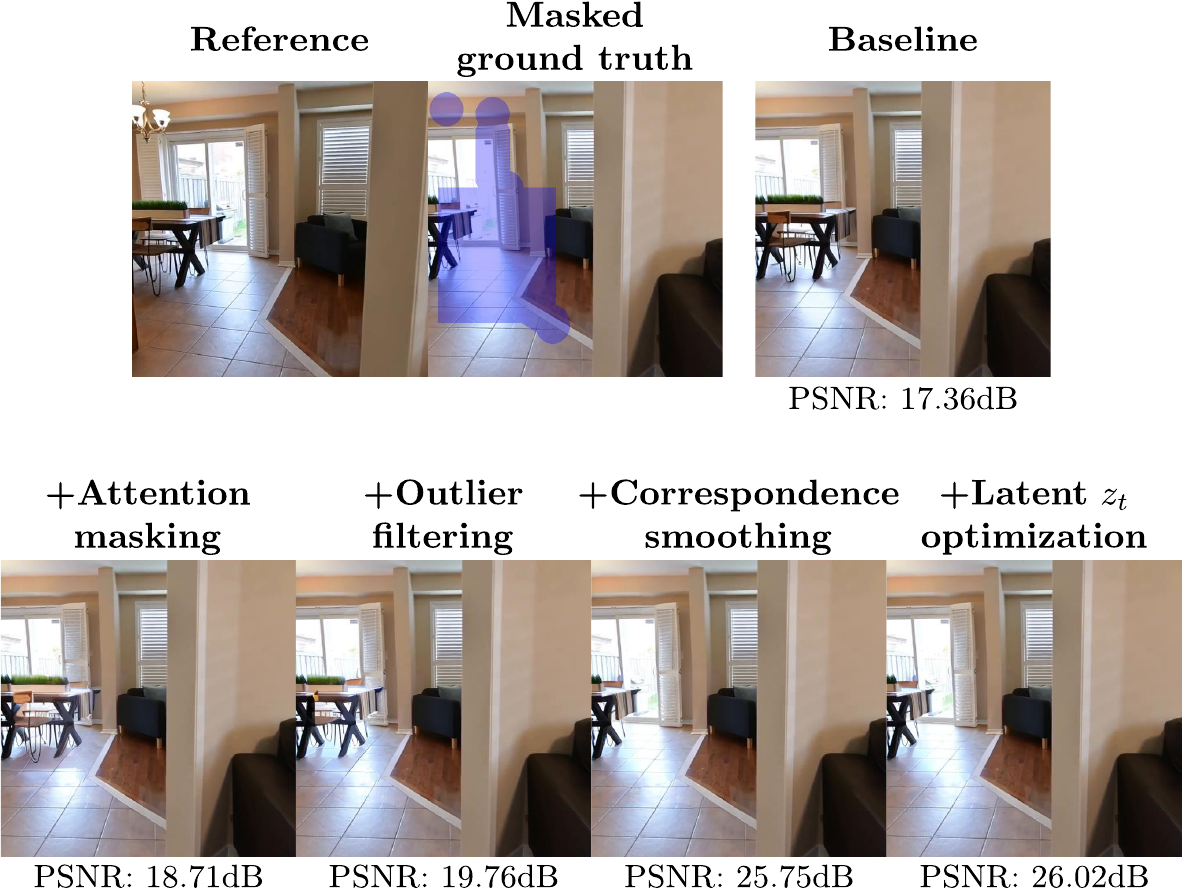}
    \caption{
    \textbf{Importance of Smoothing. }
    An example where correspondence smoothing is the pivotal component for correcting the incorrect geometry in the result of the baseline \cite{cao2024leftrefill}.
    }
    \label{fig:suppablat}
    \vspace{-3mm}
\end{figure}

%% file: Fig_Tables/Supp_Figure2.tex
\pgfplotsset{compat=1.18}
\def\axisdefaultwidth{10cm}
\def\axisdefaultheight{6cm}

\begin{figure}[t!]
\centering
\vspace{-2mm}
\scalebox{0.8}{
\begin{tikzpicture}
\begin{axis}[
axis lines=middle,
ymin=1450,
ymax=2200,
x label style={at={(current axis.right of origin)},anchor=north, below=5mm},
legend style={yshift=+0.3cm},
y label style={yshift=+0.3cm},
    xlabel=T,
    ylabel=No. of Correct Correspondences,
    xticklabel style = {anchor=mid,below=2mm},
    enlargelimits = true,
    xticklabels from table={thu2.dat}{T},xtick=data],
\addplot[blue,thick,mark=square*] table [y=All,x=X]{thu2.dat};
\addlegendentry{Ours}
\addplot[green,thick,mark=square*] table [y=NoAgg,x=X]{thu2.dat};
\addlegendentry{No acc.}]
\addplot[red,thick,mark=square*] table [y=NoCyc,x=X]{thu2.dat};
\addlegendentry{No cyc.}]
\end{axis}
\end{tikzpicture}
}

\caption{
\textbf{Numbers of correct correspondences. }
The graph illustrates the numbers of correct correspondences for three versions of \ourmethod.
``T'' denotes the timesteps of the reverse process, where inpainting progresses from $\text{T}=50$ to $0$.
``Ours'' represents our proposed method, which utilizes cyclic enhancement and estimates correspondences using aggregated attention scores across different timesteps.
``No acc.'' and ``No cyc.'' are the counterparts that exclude the accumulation of attention maps and cyclic enhancement, respectively.
}
\label{fig:numofcorr}
\end{figure}
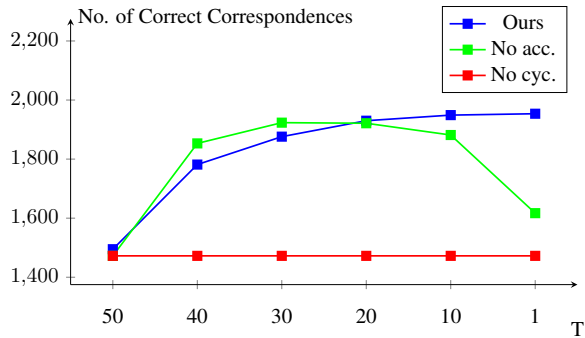

%% file: Fig_Tables/Supp_table2.tex
\begin{table}[t]
    \centering
    \scalebox{0.85}{
        \begin{tabular}{lcc}
            \toprule\midrule
            Method&Execution Time(s)&Change(s)\\
            \midrule
            Baseline&6.69&-\\
            + Attention Masking&13.77&+7.08\\
            + Outlier Filtering&14.97&+1.20\\
            + Correspondence Smoothing&15.76&+0.79\\
            + Latent $z_t$ Optimization&66.52&+50.76\\
            \midrule\bottomrule
        \end{tabular}    

        }
    \vspace{-2mm}  
    \caption{\textbf{Time analysis of key components of \ourmethod. } The execution times for the inpainting of an input were measured while incrementally enabling the key components. The baseline used in the analysis is LeftRefill.}
    \label{tab:supptable2}
\vspace{-3mm}  
\end{table}

%% file: Fig_Tables/Supp_Figure3.tex
\begin{figure}[t!]
    \centering
    \includegraphics[width=0.45\textwidth]{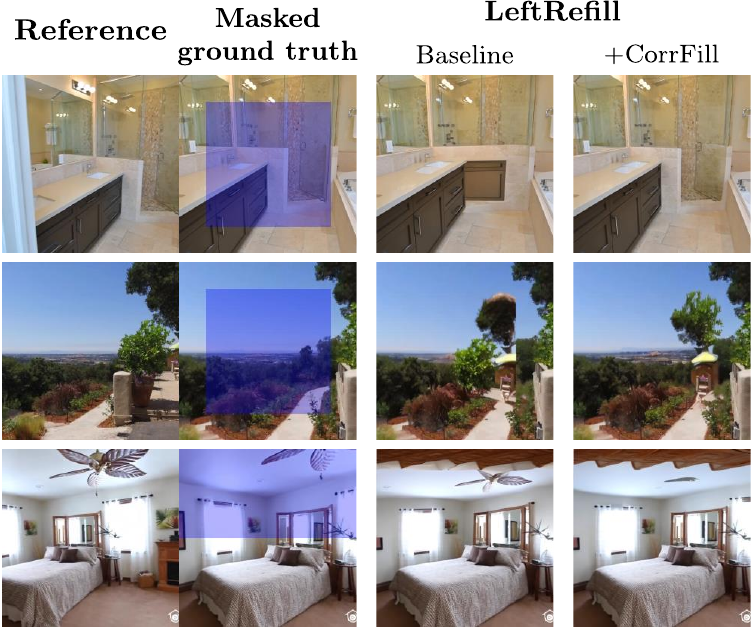}
    \caption{
    \textbf{Results with large masks. }
    The inpainting and outpainting results for the baseline method and \ourmethod~are presented. The first two rows depict the inpainting results, while the last row illustrates the outpainting results. All masks cover 50\% of the target images. \ourmethod~ cannot consistently enhance the results due to the significant degradation in the inpainting performance of the baseline method.
    }
    \label{fig:suppfail}
    \vspace{-3mm}
\end{figure}

%% file: Fig_Tables/Supp_Figure4.tex
\begin{figure}[t!]
    \centering
    \includegraphics[width=0.45\textwidth]{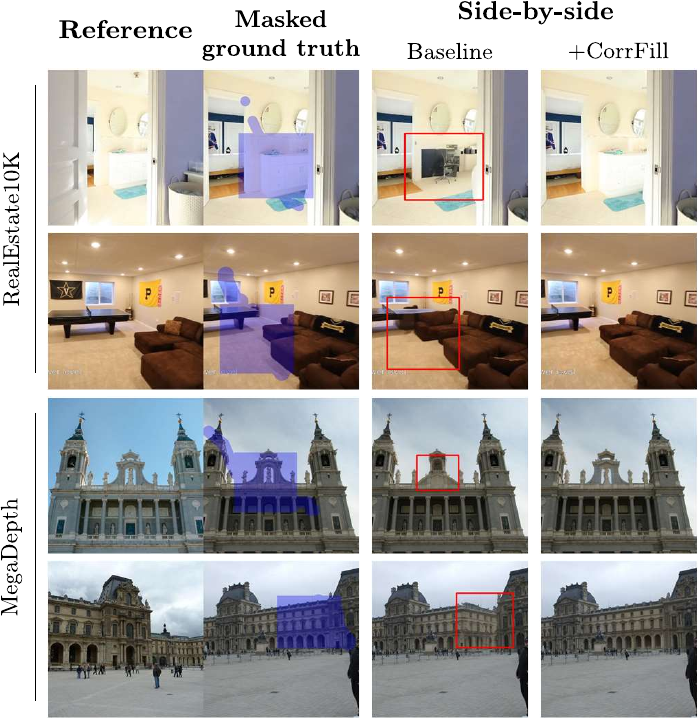}
    \caption{
    \textbf{Additional results with Side-by-side. }
    Problematic regions addressed by \ourmethod~ are highlighted within the red boxes.
    }
    \label{fig:suppexp1}
\end{figure}

%% file: Fig_Tables/Supp_Figure5.tex
\begin{figure}[t!]
    \centering
    \includegraphics[width=0.45\textwidth]{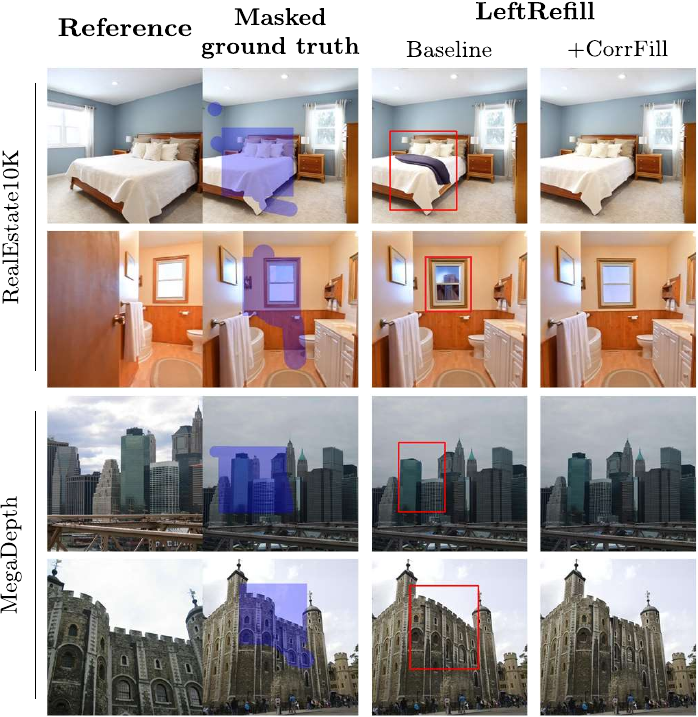}
    \caption{
    \textbf{Additional results with LeftRefill. }
    Problematic regions addressed by \ourmethod~ are highlighted within the red boxes.
    }
    \label{fig:suppexp2}
\end{figure}

%% file: main.bbl
\begin{thebibliography}{10}\itemsep=-1pt

\bibitem{balaji2023ediffi}
Yogesh Balaji, Seungjun Nah, Xun Huang, Arash Vahdat, Jiaming Song, Qinsheng Zhang, Karsten Kreis, Miika Aittala, Timo Aila, Samuli Laine, Bryan Catanzaro, Tero Karras, and Ming-Yu Liu.
\newblock ediff-i: Text-to-image diffusion models with ensemble of expert denoisers.
\newblock {\em arXiv preprint arXiv:2211.01324}, 2022.

\bibitem{cao2024leftrefill}
Chenjie Cao, Yunuo Cai, Qiaole Dong, Yikai Wang, and Yanwei Fu.
\newblock Leftrefill: Filling right canvas based on left reference through generalized text-to-image diffusion model.
\newblock In {\em CVPR}, 2024.

\bibitem{chefer2023attendnexcite}
Hila Chefer, Yuval Alaluf, Yael Vinker, Lior Wolf, and Daniel Cohen-Or.
\newblock Attend-and-excite: Attention-based semantic guidance for text-to-image diffusion models.
\newblock {\em TOG}, 2023.

\bibitem{chen2024layoutcontrol}
Minghao Chen, Iro Laina, and Andrea Vedaldi.
\newblock Training-free layout control with cross-attention guidance.
\newblock In {\em WACV}, 2024.

\bibitem{Dhariwal2021beatsGAN}
Prafulla Dhariwal and Alexander Nichol.
\newblock Diffusion models beat gans on image synthesis.
\newblock In {\em NeurIPS}, 2021.

\bibitem{epstein2024selfguidcontrol}
Dave Epstein, Allan Jabri, Ben Poole, Alexei Efros, and Aleksander Holynski.
\newblock Diffusion self-guidance for controllable image generation.
\newblock In {\em NeurIPS}, 2023.

\bibitem{feng2023structureddiffusionguidance}
Weixi Feng, Xuehai He, Tsu-Jui Fu, Varun Jampani, Arjun~Reddy Akula, Pradyumna Narayana, Sugato Basu, Xin~Eric Wang, and William~Yang Wang.
\newblock Training-free structured diffusion guidance for compositional text-to-image synthesis.
\newblock In {\em ICLR}, 2023.

\bibitem{Fu2023dreamsim}
Stephanie Fu, Netanel Tamir, Shobhita Sundaram, Lucy Chai, Richard Zhang, Tali Dekel, and Phillip Isola.
\newblock Dreamsim: Learning new dimensions of human visual similarity using synthetic data.
\newblock In {\em NeurIPS}, 2023.

\bibitem{ho2020DDPM}
Jonathan Ho, Ajay Jain, and Pieter Abbeel.
\newblock Denoising diffusion probabilistic models.
\newblock In {\em NeurIPS}, 2020.

\bibitem{kingma2022vae}
Diederik~P Kingma and Max Welling.
\newblock Auto-encoding variational bayes.
\newblock {\em arXiv preprint arXiv:1312.6114}, 2022.

\bibitem{Li2018mega}
Zhengqi Li and Noah Snavely.
\newblock Megadepth: Learning single-view depth prediction from internet photos.
\newblock In {\em CVPR}, 2018.

\bibitem{luo2023hyperfeatures}
Grace Luo, Lisa Dunlap, Dong~Huk Park, Aleksander Holynski, and Trevor Darrell.
\newblock Diffusion hyperfeatures: Searching through time and space for semantic correspondence.
\newblock In {\em NeurIPS}, 2023.

\bibitem{manukyan2023hdpainter}
Hayk Manukyan, Andranik Sargsyan, Barsegh Atanyan, Zhangyang Wang, Shant Navasardyan, and Humphrey Shi.
\newblock Hd-painter: High-resolution and prompt-faithful text-guided image inpainting with diffusion models.
\newblock {\em arXiv preprint arXiv:2312.14091}, 2023.

\bibitem{Mou2024T2I}
Chong Mou, Xintao Wang, Liangbin Xie, Yanze Wu, Jian Zhang, Zhongang Qi, and Ying Shan.
\newblock T2i-adapter: Learning adapters to dig out more controllable ability for text-to-image diffusion models.
\newblock {\em AAAI}, 2024.

\bibitem{oh2019opn}
Seoung~Wug Oh, Sungho Lee, Joon-Young Lee, and Seon~Joo Kim.
\newblock Onion-peel networks for deep video completion.
\newblock In {\em ICCV}, 2019.

\bibitem{Oquab2024DINOV2}
Maxime Oquab, Timoth{\'e}e Darcet, Th{\'e}o Moutakanni, Huy~V. Vo, Marc Szafraniec, Vasil Khalidov, Pierre Fernandez, Daniel HAZIZA, Francisco Massa, Alaaeldin El-Nouby, Mido Assran, Nicolas Ballas, Wojciech Galuba, Russell Howes, Po-Yao Huang, Shang-Wen Li, Ishan Misra, Michael Rabbat, Vasu Sharma, Gabriel Synnaeve, Hu Xu, Herve Jegou, Julien Mairal, Patrick Labatut, Armand Joulin, and Piotr Bojanowski.
\newblock {DINO}v2: Learning robust visual features without supervision.
\newblock {\em TMLR}, 2024.

\bibitem{radford2021clip}
Alec Radford, Jong~Wook Kim, Chris Hallacy, Aditya Ramesh, Gabriel Goh, Sandhini Agarwal, Girish Sastry, Amanda Askell, Pamela Mishkin, Jack Clark, Gretchen Krueger, and Ilya Sutskever.
\newblock Learning transferable visual models from natural language supervision.
\newblock In {\em ICML}, 2021.

\bibitem{rassin2024linguisticbinding}
Royi Rassin, Eran Hirsch, Daniel Glickman, Shauli Ravfogel, Yoav Goldberg, and Gal Chechik.
\newblock Linguistic binding in diffusion models: Enhancing attribute correspondence through attention map alignment.
\newblock In {\em NeurIPS}, 2023.

\bibitem{rombach2022LDM}
Robin Rombach, Andreas Blattmann, Dominik Lorenz, Patrick Esser, and Bj\"orn Ommer.
\newblock High-resolution image synthesis with latent diffusion models.
\newblock In {\em CVPR}, 2022.

\bibitem{singh2023highfidality}
Jaskirat Singh, Stephen Gould, and Liang Zheng.
\newblock High-fidelity guided image synthesis with latent diffusion models.
\newblock In {\em CVPR}, 2023.

\bibitem{song2021ddim}
Jiaming Song, Chenlin Meng, and Stefano Ermon.
\newblock Denoising diffusion implicit models.
\newblock In {\em ICLR}, 2021.

\bibitem{tang2024realfill}
Luming Tang, Nataniel Ruiz, Qinghao Chu, Yuanzhen Li, Aleksander Holynski, David~E. Jacobs, Bharath Hariharan, Yael Pritch, Neal Wadhwa, Kfir Aberman, and Michael Rubinstein.
\newblock Realfill: Reference-driven generation for authentic image completion.
\newblock {\em TOG}, 2024.

\bibitem{tang2022quadtree}
Shitao Tang, Jiahui Zhang, Siyu Zhu, and Ping Tan.
\newblock Quadtree attention for vision transformers.
\newblock In {\em ICLR}, 2022.

\bibitem{xu2023refpaint}
Dejia Xu, Xingqian Xu, Wenyan Cong, Humphrey Shi, and Zhangyang Wang.
\newblock Reference-based painterly inpainting via diffusion: Crossing the wild reference domain gap.
\newblock {\em arXiv preprint arxiv:2307.10584}, 2023.

\bibitem{yang2023PbE}
Binxin Yang, Shuyang Gu, Bo Zhang, Ting Zhang, Xuejin Chen, Xiaoyan Sun, Dong Chen, and Fang Wen.
\newblock Paint by example: Exemplar-based image editing with diffusion models.
\newblock In {\em CVPR}, 2023.

\bibitem{ye2023ipadapter}
Hu Ye, Jun Zhang, Sibo Liu, Xiao Han, and Wei Yang.
\newblock Ip-adapter: Text compatible image prompt adapter for text-to-image diffusion models.
\newblock {\em arXiv preprint arxiv:2308.06721}, 2023.

\bibitem{Zhang2024Telling}
Junyi Zhang, Charles Herrmann, Junhwa Hur, Eric Chen, Varun Jampani, Deqing Sun, and Ming-Hsuan Yang.
\newblock Telling left from right: Identifying geometry-aware semantic correspondence.
\newblock In {\em CVPR}, 2024.

\bibitem{Zhang2023zeroshot}
Junyi Zhang, Charles Herrmann, Junhwa Hur, Luisa Polania~Cabrera, Varun Jampani, Deqing Sun, and Ming-Hsuan Yang.
\newblock A tale of two features: Stable diffusion complements dino for zero-shot semantic correspondence.
\newblock In {\em NeurIPS}, 2023.

\bibitem{Zhang2023controlnet}
Lvmin Zhang, Anyi Rao, and Maneesh Agrawala.
\newblock Adding conditional control to text-to-image diffusion models.
\newblock In {\em ICCV}, 2023.

\bibitem{zhao2023geofill}
Yunhan Zhao, Connelly Barnes, Yuqian Zhou, Eli Shechtman, Sohrab Amirghodsi, and Charless Fowlkes.
\newblock Geofill: Reference-based image inpainting with better geometric understanding.
\newblock In {\em WACV}, 2023.

\bibitem{zhou2018realestate}
Tinghui Zhou, Richard Tucker, John Flynn, Graham Fyffe, and Noah Snavely.
\newblock Stereo magnification: Learning view synthesis using multiplane images.
\newblock In {\em SIGGRAPH}, 2018.

\bibitem{zhou2021transfill}
Yuqian Zhou, Connelly Barnes, Eli Shechtman, and Sohrab Amirghodsi.
\newblock Transfill: Reference-guided image inpainting by merging multiple color and spatial transformations.
\newblock In {\em CVPR}, 2021.

\end{thebibliography}


\begin{thebibliography}{1}\itemsep=-1pt

\bibitem{cao2024leftrefill}
Chenjie Cao, Yunuo Cai, Qiaole Dong, Yikai Wang, and Yanwei Fu.
\newblock Leftrefill: Filling right canvas based on left reference through generalized text-to-image diffusion model.
\newblock In {\em CVPR}, 2024.

\bibitem{shen2024gim}
Xuelun Shen, zhipeng cai, Wei Yin, Matthias M{\"u}ller, Zijun Li, Kaixuan Wang, Xiaozhi Chen, and Cheng Wang.
\newblock {GIM}: Learning generalizable image matcher from internet videos.
\newblock In {\em ICLR}, 2024.

\bibitem{von2022diffusers}
Patrick von Platen, Suraj Patil, Anton Lozhkov, Pedro Cuenca, Nathan Lambert, Kashif Rasul, Mishig Davaadorj, Dhruv Nair, Sayak Paul, William Berman, Yiyi Xu, Steven Liu, and Thomas Wolf.
\newblock Diffusers: State-of-the-art diffusion models, 2022.

\bibitem{yang2023PbE}
Binxin Yang, Shuyang Gu, Bo Zhang, Ting Zhang, Xuejin Chen, Xiaoyan Sun, Dong Chen, and Fang Wen.
\newblock Paint by example: Exemplar-based image editing with diffusion models.
\newblock In {\em CVPR}, 2023.

\bibitem{ye2023ipadapter}
Hu Ye, Jun Zhang, Sibo Liu, Xiao Han, and Wei Yang.
\newblock Ip-adapter: Text compatible image prompt adapter for text-to-image diffusion models.
\newblock {\em arXiv preprint arxiv:2308.06721}, 2023.

\end{thebibliography}
